\documentclass{article}

\PassOptionsToPackage{numbers, compress}{natbib}

     \usepackage[preprint]{neurips_2019}

\usepackage[utf8]{inputenc} 
\usepackage[T1]{fontenc}    
\usepackage{hyperref}       
\usepackage{url}            
\usepackage{booktabs}       
\usepackage{amsfonts}       
\usepackage{nicefrac}       
\usepackage{microtype}      
\usepackage[pdftex]{graphicx}

\title{Harnessing spatial MRI normalization: patch individual filter layers for CNNs}

\author{
  Fabian Eitel \\
  Humboldt Universität Berlin\\
  Berlin, 10117 \\
   \And
   Jan Philipp Albrecht \\
  Freie Universität Berlin \\
  Berlin, 14195\\
  \And
  Friedemann Paul \\
  Charité - Universitätsmedizin Berlin \\
  Berlin, 10117 \\
   \And
  Kerstin Ritter \\
  Charité - Universitätsmedizin Berlin \\
  Berlin, 10117 \\
  \texttt{kerstin.ritter@charite.de} \\
}

\begin{document}

\maketitle

\begin{abstract}

Neuroimaging studies based on magnetic resonance imaging (MRI) typically employ rigorous forms of preprocessing. Images are spatially normalized to a standard template using linear and non-linear transformations. Thus, one can assume that a patch at location \((x, y, height, width)\) contains the same brain region across the entire data set. Most analyses applied on brain MRI using convolutional neural networks (CNNs) ignore this distinction from natural images. Here, we suggest a new layer type called patch individual filter (PIF) layer, which trains higher-level filters locally as we assume that more abstract features are locally specific after spatial normalization. We evaluate PIF layers on three different tasks, namely sex classification as well as either Alzheimer's disease (AD) or multiple sclerosis (MS) detection.
We demonstrate that CNNs using PIF layers outperform their counterparts in several, especially low sample size settings.



\end{abstract}

\section{Introduction}


CNNs have been successfully applied on neuroimaging data \cite{Jo2019Review,VIEIRA2017Review}. However, several challenges have been discussed: First, sample sizes are low and public data sets typically contain no more than 1,000 patients of a specific disease. Second, MR sequences are three dimensional and can contain up to 1 million non-zero voxels, making the number of features much greater than the number of samples. Lastly, many features of the brain and neurological or psychiatric diseases are not fully understood. Even though there are guidelines for neurological assessment of diseases, these change over time (e.g. the McDonald criteria \cite{polman2011diagnostic,thompson2018diagnosis}) and only represent our current understanding of a disease.

Previous approaches using CNNs in neuroimaging tend to convert architectures which are successful on natural images to 3D MRI classifiers \cite{Guan2019comprehensive} by replacing 2D with 3D operations. In those, the special features of MRI data are typically ignored. Here, we specifically make use of the spatial homogeneity of brain MRI data. Through linear and non-linear transformations, MR images are normalized to a shared template within the MNI space such as the ICBM 152 atlas \cite{john_phd,AVANTS2008,FONOV2011313}. This ensures that a voxel at location \(L\) contains more or less the same brain region in every image and allows researchers to investigate a specific region (e.g. the hippocampus) across subjects. 
We suggest to address the spatial homogeneity by training within patches of the image and evaluate this approach in three different tasks: sex classification, AD and MS detection. Unlike patch-based approaches (see Section Related Work) we only intend to train patch-wise in higher layers. This is motivated by the idea of abstraction: whereas lower level features such as edge detectors might be globally relevant, some higher level features might be more locally relevant. 
Because higher level filters in the PIF setting train on patches which contain less information and noise than the entire image, we show here that they require fewer iterations over the training set as well as fewer samples to converge as compared to vanilla CNNs. 

\section{Related Work}
PIF layers are different from patch-based training. In patch-based training \cite{Kamnitsas2016, Ghafoorian2017, Yoo2018}, multiple patches are sampled from the dataset and fed into the same classifier regardless of the position of each patch. 
Therefore, the classifier's filters share weights between different patches. Conversely, within PIF layers, weights are only shared within a spatially restricted patch. PatchGANs \cite{li2016precomputed, isola2017image} use Markovian patches as input for a discriminator network in order to focus penalization on high-frequency structure.


PIF layers are a generalization of local convolutions as implemented in Lasagne\footnote{\url{https://lasagne.readthedocs.io/en/latest/modules/layers/local.html}} and Keras\footnote{\url{https://keras.io/layers/local/}}. Local convolutions are similar to regular convolutions but do not share weights. Local convolutions are a special case of PIF layers where \(s + p = k\), where \(s\) is the patch size, \(p\) is the padding size and \(k\) is the kernel size. Thus, the convolution kernel does not slide over the selected patch (because they are congruent). 

\section{Methods}
The heterogeneity of natural images depicting the same object requires filters to be convolved with the entire image. In a cat detection model, for example, we would expect that some higher level filters detect cat ears. In this case, it is necessary to convolve those cat ear filters with the entire image, because cat ears might be located anywhere in the image. However, when all images are spatially standardized, e.g. objects are in the same angle, viewpoint and distance and all major facial features of the cats are at the same location in each image, it would suffice if the cat ear filter searches around a small subspace of the image. This drastic form of spatial normalization is unlikely to achieve in natural images, but in neuroimaging it is the de facto standard and a major requirement for mass-univariate and multivariate pattern analysis (MVPA).
 


\begin{figure}

    \centering
    \includegraphics[width=0.82\linewidth]{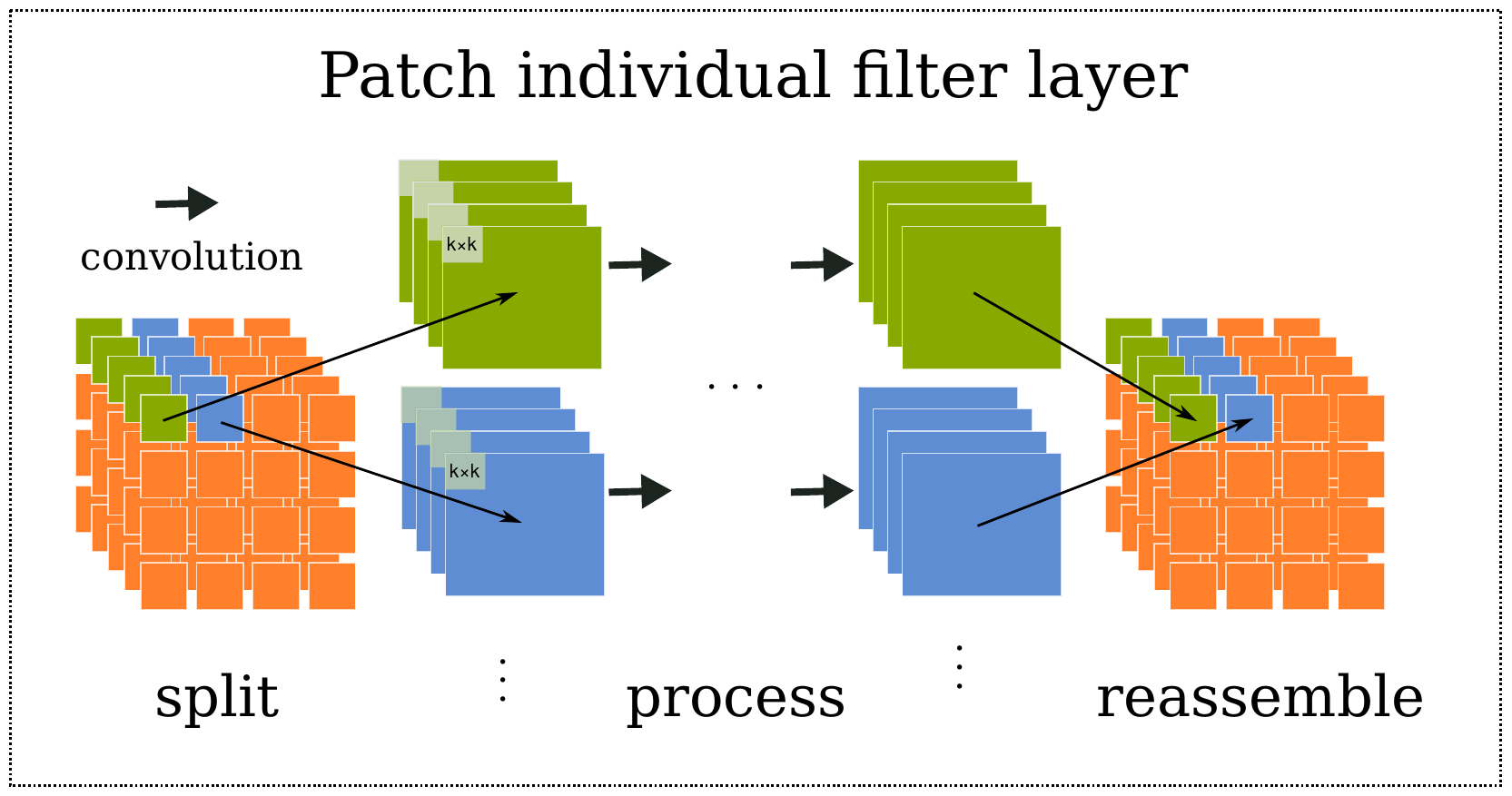}
    \caption{Depiction of a patch individual filter (PIF) layer in 2D. In this setting, inputs are 5 feature maps from a previous layer. Each feature map is being split in 16 patches and convolutions are applied patch-wise. Finally, the feature maps are reassembled in the same order.}
    \label{fig:layer}
\end{figure}

For the analysis of spatially normalized MRI data, we suggest a new CNN architecture relying on PIF layers. PIF layers consist of 3 stages: (i) split, (ii) process and (iii) reassemble. Each output feature map of the previous layer is first split (i) into patches of size \((s \times s)\). Next, the patches \(p_{ij}\) at row \(i\) and column \(j\) of all feature maps are processed (ii) with a series of local convolutions of kernel size \((k \times k)\). This is repeated for all patches. When \(s > k\), weights are shared within each patch \(p_{ij}\)  but not across patches. Lastly, all patches are reassembled (iii) in the same order as they were split. Figure \ref{fig:layer} shows an overview of the layer design. The final model consists of 4 convolutional blocks (Conv-BatchNorm-ReLU) followed by a PIF layer with a single convolutional block between split and reassemble phases.


\section{Experiments}
We evaluated PIF layers on three different datasets/tasks: 1) sex classification on a subset of the UK Biobank (F=1005, M=849), 2) AD detection on data of the Alzheimer's Disease Neuroimaging Initiative (ADNI; AD=475, HC=494) and 3) MS detection on a small private data set (MS=76, HC=71). We evaluate the performance in terms of balanced accuracy, iterations until early stopping and performance on a smaller subset. The subset contains 20\% of randomly drawn samples from the full data set.  As the number of samples for the MS data set is already small, we did not use a subset here. We compare results to a simple 5-layer CNN architecture that has shown good results on the ADNI data set and is adjusted slightly for each task. For UK Biobank and ADNI the number of parameters is smaller in the PIF architecture, whereas in MS it is larger. After finding suitable hyperparameters (learning rate, weight decay, number of filters, dropout) for each task, all experiments were repeated 10 times and averages over all repetitions are reported. Data for baseline model training was augmented using horizontal flips and translation, whereas for PIF model training only horizontal flips were used to avoid misaligned images.

\section{Results}

Table \ref{tab:results} shows the balanced accuracy and iteration in which training finished using early stopping. On the sex classification task (UK Biobank), the PIF model works almost identical on the full data set with an accuracy of almost 90\%. When using the small subset of only 20\% from the original data set, the balanced accuracy of the PIF model increased from 64.47\% to 78.11\%. On the full data set, the required iterations until early stopping almost halve, whereas on the subset they increase from 40.7 to 69.5 iterations. This is the only case where the PIF architecture has a higher number of iterations and might be due to the task-specific PIF model having a higher amount of parameters than on the other data sets.
On the AD classification task (ADNI), both baseline and PIF model perform similarly with a balanced accuracy of around 84.5\% but with a reduced set of required iterations in case of the PIF model (31.8 to 22.3). On the ADNI subset, the baseline outperformed the PIF architecture with an accuracy of 81.09\% over 76.65\% but early stopping with the PIF model occurred on average in iteration 71.9 compared to iteration 106.4 in the baseline model. 
Lastly, for MS detection balanced accuracy increased from 75.04\% to 80.92\% when using the PIF architecture and the required number of iterations decreased on average from 83.7 to 53.5. 

 \begin{table}
   \caption{Results}
   \label{tab:results}
   \centering
   \begin{tabular}{cccccc}
     \toprule
     \multicolumn{2}{c}{}
     & \multicolumn{2}{c}{\textbf{Large data set}}
     & \multicolumn{2}{c}{\textbf{Small data set}}
     \\
     Data & Model & Bal. acc. & Early stopping iter. & Bal. acc. & Early stopping iter. \\
     \midrule
     UK Biobank & Baseline-A & 89.52\% & 59.2 & 64.47\% & 40.7 \\
     UK Biobank  & PIF & 89.06\% & 33.6 & 78.11\% & 69.5 \\
     \midrule
     ADNI & Baseline-B & 84.62\% & 31.80 & 81.09\% & 106.4 \\
     ADNI & PIF & 84.43\% & 22.30 & 76.65\% & 71.9 \\
     \midrule
     MS & Baseline-C & - & - & 75.04\% & 83.7  \\
     MS & PIF & - & - & 80.92\% & 53.5 \\
     
     \bottomrule
   \end{tabular}
 \end{table}

\section{Discussion}
In this work, we have introduced a new CNN architecture relying on PIF layers to harness the established techniques of spatial normalization in neuroimaging. In multiple experiments, we have shown that PIF layers can outperform simple CNNs, especially in low sample scenarios. Further experiments are required to investigate whether the success of PIF layers is task-specific, such as the learning of regional differences in MS lesions in comparison to global atrophy in AD. 



\medskip
\small
\bibliographystyle{unsrtnat}
\bibliography{main}

\begin{thebibliography}{13}
\providecommand{\natexlab}[1]{#1}
\providecommand{\url}[1]{\texttt{#1}}
\expandafter\ifx\csname urlstyle\endcsname\relax
  \providecommand{\doi}[1]{doi: #1}\else
  \providecommand{\doi}{doi: \begingroup \urlstyle{rm}\Url}\fi

\bibitem[Jo et~al.(2019)Jo, Nho, and Saykin]{Jo2019Review}
Taeho Jo, Kwangsik Nho, and Andrew~J. Saykin.
\newblock Deep learning in alzheimer's disease: Diagnostic classification and
  prognostic prediction using neuroimaging data.
\newblock \emph{Frontiers in Aging Neuroscience}, 11:\penalty0 220, 2019.
\newblock ISSN 1663-4365.
\newblock \doi{10.3389/fnagi.2019.00220}.
\newblock URL
  \url{https://www.frontiersin.org/article/10.3389/fnagi.2019.00220}.

\bibitem[Vieira et~al.(2017)Vieira, Pinaya, and Mechelli]{VIEIRA2017Review}
Sandra Vieira, Walter~H.L. Pinaya, and Andrea Mechelli.
\newblock Using deep learning to investigate the neuroimaging correlates of
  psychiatric and neurological disorders: Methods and applications.
\newblock \emph{Neuroscience \& Biobehavioral Reviews}, 74:\penalty0 58 -- 75,
  2017.
\newblock ISSN 0149-7634.
\newblock \doi{https://doi.org/10.1016/j.neubiorev.2017.01.002}.
\newblock URL
  \url{http://www.sciencedirect.com/science/article/pii/S0149763416305176}.

\bibitem[Polman et~al.(2011)Polman, Reingold, Banwell, Clanet, Cohen, Filippi,
  Fujihara, Havrdova, Hutchinson, Kappos, et~al.]{polman2011diagnostic}
Chris~H Polman, Stephen~C Reingold, Brenda Banwell, Michel Clanet, Jeffrey~A
  Cohen, Massimo Filippi, Kazuo Fujihara, Eva Havrdova, Michael Hutchinson,
  Ludwig Kappos, et~al.
\newblock Diagnostic criteria for multiple sclerosis: 2010 revisions to the
  mcdonald criteria.
\newblock \emph{Annals of neurology}, 69\penalty0 (2):\penalty0 292--302, 2011.

\bibitem[Thompson et~al.(2018)Thompson, Banwell, Barkhof, Carroll, Coetzee,
  Comi, Correale, Fazekas, Filippi, Freedman, et~al.]{thompson2018diagnosis}
Alan~J Thompson, Brenda~L Banwell, Frederik Barkhof, William~M Carroll, Timothy
  Coetzee, Giancarlo Comi, Jorge Correale, Franz Fazekas, Massimo Filippi,
  Mark~S Freedman, et~al.
\newblock Diagnosis of multiple sclerosis: 2017 revisions of the mcdonald
  criteria.
\newblock \emph{The Lancet Neurology}, 17\penalty0 (2):\penalty0 162--173,
  2018.

\bibitem[Guan et~al.(2019)Guan, Kumar, Fung, Wu, and
  Fiterau]{Guan2019comprehensive}
Ziqiang Guan, Ritesh Kumar, Yi~Ren Fung, Yeahuay Wu, and Madalina Fiterau.
\newblock A comprehensive study of alzheimer's disease classification using
  convolutional neural networks.
\newblock \emph{CoRR}, abs/1904.07950, 2019.
\newblock URL \url{http://arxiv.org/abs/1904.07950}.

\bibitem[Ashburner(2000)]{john_phd}
J.~Ashburner.
\newblock \emph{Computational Neuroanatomy}.
\newblock PhD thesis, University College London, 2000.
\newblock URL \url{/spm/doc/theses/john/}.

\bibitem[Avants et~al.(2008)Avants, Epstein, Grossman, and Gee]{AVANTS2008}
B.B. Avants, C.L. Epstein, M.~Grossman, and J.C. Gee.
\newblock Symmetric diffeomorphic image registration with cross-correlation:
  Evaluating automated labeling of elderly and neurodegenerative brain.
\newblock \emph{Medical Image Analysis}, 12\penalty0 (1):\penalty0 26 -- 41,
  2008.
\newblock ISSN 1361-8415.
\newblock \doi{https://doi.org/10.1016/j.media.2007.06.004}.
\newblock URL
  \url{http://www.sciencedirect.com/science/article/pii/S1361841507000606}.
\newblock Special Issue on The Third International Workshop on Biomedical Image
  Registration – WBIR 2006.

\bibitem[Fonov et~al.(2011)Fonov, Evans, Botteron, Almli, McKinstry, and
  Collins]{FONOV2011313}
Vladimir Fonov, Alan~C. Evans, Kelly Botteron, C.~Robert Almli, Robert~C.
  McKinstry, and D.~Louis Collins.
\newblock Unbiased average age-appropriate atlases for pediatric studies.
\newblock \emph{NeuroImage}, 54\penalty0 (1):\penalty0 313 -- 327, 2011.
\newblock ISSN 1053-8119.
\newblock \doi{https://doi.org/10.1016/j.neuroimage.2010.07.033}.
\newblock URL
  \url{http://www.sciencedirect.com/science/article/pii/S1053811910010062}.

\bibitem[Kamnitsas et~al.(2016)Kamnitsas, Ferrante, Parisot, Ledig, Nori,
  Criminisi, Rueckert, and Glocker]{Kamnitsas2016}
Konstantinos Kamnitsas, Enzo Ferrante, Sarah Parisot, Christian Ledig,
  Aditya~V. Nori, Antonio Criminisi, Daniel Rueckert, and Ben Glocker.
\newblock Deepmedic for brain tumor segmentation.
\newblock In Alessandro Crimi, Bjoern Menze, Oskar Maier, Mauricio Reyes,
  Stefan Winzeck, and Heinz Handels, editors, \emph{Brainlesion: Glioma,
  Multiple Sclerosis, Stroke and Traumatic Brain Injuries}, pages 138--149,
  Cham, 2016. Springer International Publishing.
\newblock ISBN 978-3-319-55524-9.

\bibitem[Ghafoorian et~al.(2017)Ghafoorian, Karssemeijer, Heskes, Bergkamp,
  Wissink, Obels, Keizer, de~Leeuw, van Ginneken, Marchiori, and
  Platel]{Ghafoorian2017}
Mohsen Ghafoorian, Nico Karssemeijer, Tom Heskes, Mayra Bergkamp, Joost
  Wissink, Jiri Obels, Karlijn Keizer, Frank-Erik de~Leeuw, Bram van Ginneken,
  Elena Marchiori, and Bram Platel.
\newblock {Deep multi-scale location-aware 3D convolutional neural networks for
  automated detection of lacunes of presumed vascular origin.}
\newblock \emph{NeuroImage. Clinical}, 14:\penalty0 391--399, 2017.
\newblock ISSN 2213-1582.
\newblock \doi{10.1016/j.nicl.2017.01.033}.
\newblock URL \url{http://www.ncbi.nlm.nih.gov/pubmed/28271039
  http://www.pubmedcentral.nih.gov/articlerender.fcgi?artid=PMC5322213}.

\bibitem[Yoo et~al.(2018)Yoo, Tang, Brosch, Li, Kolind, Vavasour, Rauscher,
  MacKay, Traboulsee, and Tam]{Yoo2018}
Youngjin Yoo, Lisa~Y.W. Tang, Tom Brosch, David~K.B. Li, Shannon Kolind, Irene
  Vavasour, Alexander Rauscher, Alex~L. MacKay, Anthony Traboulsee, and
  Roger~C. Tam.
\newblock {Deep learning of joint myelin and T1w MRI features in
  normal-appearing brain tissue to distinguish between multiple sclerosis
  patients and healthy controls}.
\newblock \emph{NeuroImage: Clinical}, 17:\penalty0 169--178, 2018.
\newblock ISSN 22131582.
\newblock \doi{10.1016/j.nicl.2017.10.015}.
\newblock URL
  \url{http://linkinghub.elsevier.com/retrieve/pii/S2213158217302553}.

\bibitem[Li and Wand(2016)]{li2016precomputed}
Chuan Li and Michael Wand.
\newblock Precomputed real-time texture synthesis with markovian generative
  adversarial networks.
\newblock In \emph{European Conference on Computer Vision}, pages 702--716.
  Springer, 2016.

\bibitem[Isola et~al.(2017)Isola, Zhu, Zhou, and Efros]{isola2017image}
Phillip Isola, Jun-Yan Zhu, Tinghui Zhou, and Alexei~A Efros.
\newblock Image-to-image translation with conditional adversarial networks.
\newblock In \emph{Proceedings of the IEEE conference on computer vision and
  pattern recognition}, pages 1125--1134, 2017.

\end{thebibliography}

\end{document}